\documentclass[pdflatex,sn-mathphys-num]{sn-jnl}

\usepackage{graphicx}%
\usepackage{multirow}%
\usepackage{amsmath,amssymb,amsfonts}%
\usepackage{amsthm}%
\usepackage{mathrsfs}%
\usepackage[title]{appendix}%
\usepackage{xcolor}%
\usepackage{textcomp}%
\usepackage{manyfoot}%
\usepackage{algorithm}%
\usepackage{pifont} 
\usepackage{algorithmicx}%
\usepackage{algpseudocode}%
\usepackage{listings}%
\usepackage{mathtools}
\usepackage{siunitx}
\usepackage{placeins}
\usepackage{subcaption}

\usepackage{tikz}
\usetikzlibrary{mindmap,trees,arrows.meta,positioning}
\usepackage{rotating}          
\usepackage{adjustbox}         
\usepackage{booktabs}          
\usepackage{threeparttable}    
\usepackage{makecell}          
\usepackage{tabularx,array,float}
\newcolumntype{L}[1]{>{\raggedright\arraybackslash}p{#1}}
\newcolumntype{C}[1]{>{\centering\arraybackslash}p{#1}}

\usepackage{pifont}
\newcommand{\cmark}{\ding{51}}
\newcommand{\xmark}{\ding{55}}
\newcommand{\pmark}{\raisebox{0.2ex}{\tiny$\triangle$}}

\usepackage[utf8]{inputenc}

\DeclareUnicodeCharacter{2009}{\,}

\theoremstyle{thmstyleone}%
\theoremstyle{thmstyletwo}%

\theoremstyle{thmstylethree}%

\usepackage{array}
\newcolumntype{L}[1]{>{\raggedright\arraybackslash}p{#1}}
\newcolumntype{C}[1]{>{\centering\arraybackslash}p{#1}}

\begin{document}

\title[Article Title]{From Graphs to Gates: DNS-HyXNet, A Lightweight and Deployable Sequential Model for Real-Time DNS Tunnel Detection}

\author[1]{\normalsize \fnm{Faraz} \sur{Ali}}
\author[1,2]{\normalsize \fnm{Muhammad} \sur{Afaq}}
\author[1,2]{\normalsize \fnm{Mahmood} \sur{Niazi}}
\author*[1,3]{\normalsize \fnm{Muzammil} \sur{Behzad}}\email{muzammil.behzad@kfupm.edu.sa}



\affil[1]{\normalsize \orgname{King Fahd University of Petroleum and Minerals}, \orgaddress{\country{Saudi Arabia}}}
\affil[2]{\normalsize \orgname{Interdisciplinary Research Center for Intelligent Secure Systems}, \orgaddress{\country{Saudi Arabia}}}
\affil[3]{\normalsize \orgname{SDAIA-KFUPM Joint Research Center for Artificial Intelligence}, \orgaddress{\country{Saudi Arabia}}}

\abstract{
\textbf{Purpose:} Domain Name System (DNS) tunneling remains a covert channel for data exfiltration and command-and-control communication.  
Graph-based models such as GraphTunnel attain strong accuracy but incur latency and overhead from recursive parsing and graph construction, limiting real-time use.  
This study presents DNS-HyXNet, a lightweight extended Long Short-Term Memory (xLSTM) Hybrid framework for efficient sequence-based DNS tunnel detection.

\textbf{Methods:}
DNS-HyXNet integrates tokenized domain embeddings with normalized numeric DNS features and processes them through a two-layer xLSTM network that learns temporal dependencies directly from packet sequences.  
The model removes graph reconstruction entirely, enabling single-stage multi-class classification.  
Training and evaluation on two public benchmarks ensured fair comparison, with hyperparameters tuned for low memory use and fast inference.

\textbf{Results:}
Across all experimental splits of the DNS-Tunnel-Datasets, DNS-HyXNet achieved up to 99.99\% accuracy, with macro-averaged precision, recall, and F1-scores remaining above 99.96\%.
Its per-sample detection latency of 0.041~ms demonstrates scalability and real-time readiness.

\textbf{Conclusions:}
Sequential modeling with xLSTM replaces recursive graph generation, cutting computational cost while preserving top-tier accuracy.  
The framework enables real-time DNS tunnel detection on commodity hardware, offering a deployable and energy-efficient alternative to graph-based methods.}

\keywords{Artificial Intelligence, Deep Learning, DNS Tunneling Detection, DNS-HyXNet, xLSTM, Real-Time Network Security}

\maketitle

\section{Introduction}\label{sec1}

The Domain Name System (DNS) is an indispensable component of the Internet infrastructure, responsible for translating human-readable domain names into numerical IP addresses. Its ubiquity and the fact that port 53 is almost universally permitted through firewalls make DNS an attractive target for adversaries seeking covert communication channels. Recent surveys indicate that more than two-thirds of global enterprises suffered DNS-based attacks within the past year, with tunneling and data exfiltration ranking among the most frequent causes of compromise \cite{coker202172, EfficientIP_IDC2022}.  
DNS tunneling enables attackers to encapsulate arbitrary payloads inside legitimate DNS queries and responses, thereby bypassing intrusion-prevention systems and network address translation policies \cite{wang2021comprehensive, adiwal2023dns}. The technique has been adopted in numerous malware families from Feederbot and Morto to advanced persistent threat (APT) campaigns such as APT32 and APT34 which use DNS to exchange command-and-control (C2) messages and to exfiltrate sensitive data \cite{salat2023dns, luo2020towards, abualghanam2023realtime}. 

Early tunnel detection strategies relied on rule-based or statistical heuristics, such as fixed thresholds for domain length, character entropy, and query frequency \cite{sheridan2015detection, ellens2013flow, bai2021refined}.  
While effective for known signatures, these approaches struggle to generalize to encrypted or obfuscated tunnels that manipulate DNS payloads dynamically.  
To improve generalization, researchers turned to machine-learning classifiers using lexical, syntactic, and behavioral features of fully qualified domain names (FQDNs) and resource records. Support-vector machines (SVMs), random forests, and isolation forests have been applied with considerable success for binary or multi-label tunnel classification \cite{ishikura2021dns, d2022dns, liang2023fecc}.  
However, feature-engineered ML models depend heavily on manual attribute selection and require frequent retraining when attackers alter encoding schemes or protocol behavior \cite{salat2023dns, liang2023fecc, abualghanam2023realtime}.  

More recently, deep-learning paradigms have demonstrated their potential to automate feature extraction and improve accuracy.  
Convolutional neural networks (CNNs) \cite{liang2023fecc} and recurrent neural networks (RNNs) \cite{palau2020dns} have been explored to capture latent lexical or temporal structures within DNS traffic.  
Hybrid frameworks such as METC \cite{zuo2025metc} and FECC \cite{liang2023fecc} achieved promising results by integrating multiple learning paradigms, yet their inference latency and hardware demands remain high.  
At the same time, explainable and interpretable detection mechanisms have gained attention, as seen in the Rule-Based eXplainable Autoencoder (ReXAE) \cite{debernardi2025rule}, which combines transparency with deep representation learning.  

The most influential recent work is Gao et al.’s GraphTunnel \cite{gao2024graphtunnel}, a graph neural network (GNN) framework that models recursive DNS-resolution chains as node-edge graphs.  
By learning contextual relationships among domain queries, GraphTunnel achieved robust detection of both known and unseen tunneling tools.
Nevertheless, the approach incurs significant computational and temporal overhead from graph construction and recursive traversal, making it more suitable for offline forensic analysis than for real-time intrusion prevention.  
Its two-stage pipeline graph classification followed by CNN-based tunnel tool identification further increases latency and complexity, limiting deployability on edge gateways or high-throughput enterprise resolvers. 

In operational environments, DNS tunnel detection must perform in real time at the network perimeter, where milliseconds determine whether sensitive data are exfiltrated.  
Recursive graph modeling, as employed by GNN based solutions, requires observation of the entire resolution chain and therefore cannot respond before compromise.  
Moreover, large scale corporate networks require lightweight and memory-efficient algorithms that maintain high accuracy without GPU dependence or costly preprocessing pipelines \cite{ozery2024information, adiwal2023dns}.  

To address these limitations, we propose DNS-HyXNet, an extended Long Short-Term Memory (xLSTM) driven hybrid model that captures sequential and statistical characteristics of DNS traffic without explicit graph construction.  
The framework processes tokenized lexical embeddings of domain queries alongside normalized numeric attributes enabling the network to learn temporal correlations between benign and tunneled exchanges.  
Unlike graph-centric architectures, DNS-HyXNet directly models packet sequences, thus supporting real-time inference while preserving contextual awareness across query-response events.  
The xLSTM’s exponential gating mechanism enhances its memory retention and gradient stability compared with standard LSTM or GRU designs, enabling accurate discrimination of low-throughput tunnels that typically evade threshold or rule-based detectors \cite{luo2020towards, abualghanam2023realtime}. 

Comprehensive evaluation on the public DNS-Tunnel-Datasets \cite{gao2024dataset}, the same corpus used by GraphTunnel demonstrates that DNS-HyXNet achieves near-perfect detection accuracy for both known and unseen tunneling tools while reducing inference latency by an order of magnitude.  
The proposed single-stage multi-class classifier eliminates the need for a secondary identification module, substantially lowering computational overhead.  
Its lightweight architecture allows seamless deployment at DNS gateways, enterprise firewalls, and SOC edge appliances, representing a practical step toward adaptive, real-time DNS threat defense.

The remainder of this article is organized as follows: Section \ref{sec2} reviews prior research in DNS tunneling detection and deep sequence modeling; Section \ref{sec3} details the DNS-HyXNet architecture and data preprocessing, Section \ref{sec4} presents the experimental setup and results, Section \ref{sec5} discusses practical implications and limitations, and Section \ref{sec6} concludes the paper with future directions.

\section{Literature Review}\label{sec2}
This section reviews the evolution of DNS tunneling detection techniques, highlighting the strengths and limitations of prior approaches and identifying the gaps that motivate the development of DNS-HyXNet.
\subsection{Why DNS tunneling remains difficult to detect}
Despite two decades of research, DNS tunneling continues to evade enterprise detection systems because the protocol itself is inherently trusted and its queries are rarely blocked or inspected.  
DNS operates on UDP port 53, a service universally permitted by firewalls, and this trust is exploited by attackers who encapsulate command-and-control (C2) or exfiltration payloads inside legitimate DNS traffic \cite{wang2021comprehensive, coker202172}.  
Studies reveal that DNS traffic is now among the most abused channels for covert communication, with over 70\% of organizations reporting DNS-based incidents \cite{EfficientIP_IDC2022}.  
The challenge persists because tunneling traffic often mimics benign recursive lookups, exhibits low throughput, and dynamically changes domain names, rendering static inspection rules obsolete \cite{adiwal2023dns, salat2023dns, luo2020towards}.

Research efforts have evolved from early rule and signature-based systems to statistical modeling, classical machine learning (ML), and deep learning (DL) frameworks, culminating in recent structure-aware and hybrid designs.  
However, three unresolved gaps remain consistent across the literature:  
(i) robustness against previously unseen tunneling tools and wildcard DNS responses,  
(ii) operational readiness for real-time, low-memory, high-throughput deployment, and  
(iii) multi-class attribution beyond binary malicious/benign classification \cite{wang2022krtunnel, gao2024graphtunnel}.  
Graph-based models improved generalization but incur graph-construction latency and heavy memory use, limiting inline applicability \cite{bai2021refined, gao2024graphtunnel}.  
These issues motivate research into sequential models such as DNS-HyXNet that retain temporal awareness without recursive parsing.

\subsection{Evolution of detection paradigms}
Over the years, DNS tunnel detection has progressed through several distinct methodological phases, each introducing new capabilities while exposing new limitations. The major paradigms that shaped this evolution are outlined below.
\subsubsection{Rule-based and signature methods}
Initial detection frameworks relied on hand-crafted rules and static indicators such as specific keyword tokens, domain label formats, or fixed thresholds for length and entropy \cite{sheridan2015detection, adiwal2023dns}.  
Systems like SNORT-based DID \cite{adiwal2023dns} and DNSGuard heuristics embedded expert rules to flag abnormal TTL values or suspicious resource-record (RR) patterns.  
Although computationally light and easy to interpret, these methods rapidly degrade when adversaries randomize encodings, employ subdomain fluxing, or tunnel over encrypted channels such as DoH and DoT \cite{ding2021encrypt, zuo2025metc}.  
Rule-based mechanisms therefore remain useful for baseline monitoring but are insufficient for adaptive threats.

\subsubsection{Statistical and threshold-driven approaches}
To reduce dependency on fixed signatures, several works introduced probabilistic or information-theoretic features.  
Leijenhorst et~al. quantified tunnel throughput and latency in early feasibility analyses \cite{van2008viability}, while Aiello et~al. performed a comprehensive performance assessment across tools such as DNSCat, Iodine, and Dns2TCP \cite{merlo2011comparative}.  
Subsequent models used entropy, frequency distributions, or inter-arrival-time deviations to detect anomalous flows \cite{ozery2024information, ishikura2021dns}.  
These designs generalized better than fixed rules but required carefully tuned thresholds and still struggled with wildcard responses and encrypted DNS streams \cite{bai2021refined}.

\subsubsection{Classical machine-learning models}
Machine learning introduced data-driven classification on engineered features such as packet length, query type ratios, subdomain entropy, and record counts.  
Random Forests and Support-Vector Machines (SVMs) achieved strong accuracy on curated datasets \cite{yang2020detecting, shafieian2023multi}, yet their reliance on manual feature extraction limited portability across networks.  
Isolation-Forest detectors such as KRTunnel extended this concept to Android and mobile environments, demonstrating light anomaly detection on resource-constrained endpoints \cite{wang2022krtunnel}.  
However, these models lacked temporal context and performed poorly when faced with unseen encoding schemes or adaptive query timing \cite{merlo2011comparative, ozery2024information}.

\subsubsection{Deep learning and representation learning}
Deep architectures marked a turning point by automatically learning features from raw inputs.  
Convolutional Neural Networks (CNNs) converted DNS flows into image like matrices (DNS-Images) \cite{d2022dns} or feature clusters (FECC) \cite{liang2023fecc}, achieving over 98\% accuracy on known tunnels.  
Recurrent networks captured sequential dependencies within query streams, Palau et~al. employed LSTMs on lexical sequences to model domain morphologies \cite{palau2020dns}.  
Hybrid deep frameworks followed: METC fused CNN and RNN encoders for DoH traffic \cite{zuo2025metc}, while DLAZE \cite{sharma2025dlaze} and Real-Time ML detectors \cite{abualghanam2023realtime} leveraged ensemble or hybrid feature selection.  
Although these models outperform traditional ML in detection accuracy, most are computationally expensive and not suited for inline processing on DNS resolvers or IoT gateways.

\subsubsection{Structure-aware and graph-based models}
Recognizing that DNS queries inherently form recursive graphs, recent research integrated graph learning.  
Bai et~al. used regression analysis to identify hybrid protocol traffic within tunnels \cite{bai2021refined}.  
Gao et~al. introduced GraphTunnel, a robust graph-neural-network (GNN) framework that models recursive resolution paths as nodes and edges, aggregates contextual embeddings via GraphSAGE, and classifies tunnel families through a CNN layer \cite{gao2024graphtunnel}.  
GraphTunnel achieved near-perfect F1 on known and unseen tunnels, outperforming CNN baselines such as FECC and DNS-Images \cite{liang2023fecc, d2022dns}.  
However, its graph-construction latency, high memory footprint, and two-stage pipeline make it impractical for real-time deployment.  
Explainable variants such as the Rule-Based eXplainable Autoencoder (ReXAE) \cite{debernardi2025rule} improved interpretability but remained limited to offline analysis.

\subsubsection{Emerging efficiency-oriented sequence models}
Given these trade-offs, researchers are exploring sequence models that learn temporal dependencies directly from packet streams without building global resolution graphs.  
xLSTM style architectures \cite{luo2020towards, abualghanam2023realtime} combine recurrent gating with exponential memory decay, allowing long-range temporal modeling at low computational cost.  
Such models can integrate tokenized categorical embeddings (domain labels, query types) with normalized numerical attributes (entropy, TTL, packet size), enabling unified real-time detection.  
The proposed DNS-HyXNet extends this paradigm, offering single-stage multi-class classification with millisecond inference latency bridging the gap between high-accuracy deep models and operational efficiency.

\subsection{Comparative analysis of representative works}
To situate DNS-HyXNet within the broader research landscape, we compare it against prominent DNS tunneling detection approaches reported in the literature. Tables~\ref{tab:compA} and~\ref{tab:compB} summarize methodological differences, reported performance, and operational capabilities across these systems.
\begin{table}[!t] 
\centering
\caption{Comparison of DNS tunneling detection methods in terms of methodology and accuracy}
\label{tab:compA}
\footnotesize
\renewcommand{\arraystretch}{1.08}
\setlength{\tabcolsep}{3.5pt}

\begin{tabularx}{\textwidth}{
  L{2.6cm}  
  C{0.7cm}  
  L{3.3cm}  
  C{1.4cm}  
  >{\raggedright\arraybackslash}X  
}
\toprule
Study / Model & Year & Methodology & Accuracy & Notes \\
\midrule
Sheridan \& Keane~\cite{sheridan2015detection} & 2015 & Signature / Rule & 90--95\% & Simple thresholds; brittle under encryption; poor generalization. \\
Adiwal et al.~\cite{adiwal2023dns} & 2023 & Rule-based SNORT & -- & Requires rule curation; weak on unseen variants. \\
Leijenhorst et al. & 2008 & Empirical benchmark & -- & Quantified DNScat throughput; early baselines. \\
Aiello et al. & 2011 & Comparative tool analysis & -- & Early performance taxonomy of tunnel tools. \\
Yang et al.~\cite{yang2020detecting} & 2020 & RF (session behavior) & 96\% & Poor cross-dataset generalization; limited streaming evaluation. \\
Wang et al.~\cite{wang2022krtunnel} & 2022 & Isolation Forest (mobile) & 98\% & Lightweight; limited tool diversity; Android focus. \\
D’Angelo et al.~\cite{d2022dns} & 2022 & CNN (DNS-images) & 99.7\% & Strong on known tunnels; weaker OOD robustness. \\
Liang et al.~\cite{liang2023fecc} & 2023 & CNN + clustering (FECC) & 95-99\% & High accuracy; higher compute cost. \\
Palau et al.~\cite{palau2020dns} & 2020 & LSTM (lexical) & 97\% & No streaming evaluation; focuses on domain-string patterns. \\
Ishikura et al.~\cite{ishikura2021dns} & 2021 & Cache-aware features & 99\% & Feature-free extraction; offline only (no inline prevention). \\
\bottomrule
\end{tabularx}
\end{table}

\begin{table*}[!b]
\centering
\caption{Comparison of DNS tunneling detection methods in terms of robustness and efficiency\newline
\footnotesize
\textit{Columns:} 
U-Tunnel = Unknown-Tunnel robustness; 
W-DNS = Wildcard-DNS handling; 
RT = Real-Time capability; 
Mem. = Memory-Efficient; 
Tool ID = Tool attribution capability.
\newline
\textit{Symbols:} 
\cmark = supported, 
\xmark = not supported, 
\pmark = partially or unclearly reported.}
\label{tab:compB}
\scriptsize
\setlength{\tabcolsep}{4pt}
\resizebox{\textwidth}{!}{%
\begin{tabular}{lccccccc}
\toprule
Study / Model & U-Tunnel & W-DNS & RT & Mem. & Tool ID & Accuracy & Notes\\
\midrule
Bai et al.~\cite{bai2021refined}        & \cmark & \cmark & \xmark & \xmark & \xmark & 92\%   & Multi-protocol regression.\\
Abualghanam et al.~\cite{abualghanam2023realtime} & \cmark & \pmark & \cmark & \pmark & \xmark & 98.3\%& Hybrid ensemble.\\
Salat et al.~\cite{salat2023dns}        & \cmark & \cmark & \pmark & \pmark & \xmark & 99\%   & Cloud ML; heavy resource use.\\
De Bernardi et al.~\cite{debernardi2025rule} & \cmark & \cmark & \xmark & \pmark & \xmark & 97\%   & Explainable, offline.\\
Gao et al.~\cite{gao2024graphtunnel}    & \cmark & \cmark & \pmark & \xmark & \cmark & 99.78\% & Graph parsing latency.\\
Luo et al.~\cite{luo2020towards}        & \cmark & \cmark & \cmark & \cmark & \xmark & 99\%   & First A/AAAA RR study.\\
Ozery et al.~\cite{ozery2024information} & \pmark & \xmark & \cmark & \cmark & \xmark & 96\%   & Real-time heavy hitters.\\
DNS-HyXNet (ours)                     & \cmark & \cmark & \cmark & \cmark & \cmark & 99.99\% & Single-stage, low-memory.\\
\bottomrule
\end{tabular}%
}
\end{table*}

\subsection{Synthesis and research gaps}
The evolution from static rules to deep learning has markedly improved tunnel detection accuracy, yet operational limitations persist.  
Rule-based and statistical detectors are interpretable and fast but fail against encrypted or polymorphic tunnels.  
Classical ML methods increase adaptability but depend on engineered features that cannot capture temporal dynamics.  
Deep CNN and RNN frameworks outperform earlier models on benchmark datasets but sacrifice deployability due to GPU demand and inference latency.  
Graph-based models such as GraphTunnel further enhance robustness to unknown tunnels and wildcard DNS by encoding recursive relationships, yet their graph-construction cost precludes inline use in high-speed resolvers.  
Explainable variants improve transparency but not performance.

Therefore, three gaps emerge across the literature:  
(i) a need for single-stage models capable of detecting and attributing tunnel types simultaneously
(ii) a demand for streaming-first, memory-efficient frameworks that avoid recursive graph parsing and  
(iii) the absence of unified evaluations that measure latency and energy consumption alongside accuracy.  
DNS-HyXNet directly addresses these challenges by learning temporal patterns through xLSTM gating while integrating lexical and statistical cues in a unified sequential model.  
Its lightweight recurrent design achieves the robustness of graph models and the speed of threshold detectors, aligning academic precision with operational practicality.

\section{The Proposed DNS-HyXNet Method}\label{sec3}
This section details the methodological foundations of DNS-HyXNet, outlining the principles that guided its design and the technical components that enable real-time, graph-free DNS tunnel detection. We begin by stating the core design objectives that shaped the architecture.
\subsection{Design objectives}
We design DNS-HyXNet to address three gaps that persist across prior art, (i) real-time operation without the latency and state blow-up of recursive graph construction, (ii) single-stage multi-class output (maliciousness and tool family) to eliminate multi-model pipelines, and (iii) small memory footprint so inference can run on commodity gateways and enterprise resolvers. To this end DNS-HyXNet is strictly graph-free and sequence-first: temporal regularities in DNS behavior are modeled directly with a compact xLSTM backbone, and a single dense head performs unified classification.

\subsubsection{Addressing known research gaps}
The design choices in DNS-HyXNet directly respond to limitations identified in the literature:
\begin{itemize}
  \item Graph-free, streaming-first: Eliminates the latency and state overhead of recursive graph construction, enabling real-time inference.
  \item Single-stage multi-class: Produces both maliciousness and tool attribution in one forward pass, reducing cascaded model failures.
  \item Robust tokenization: Bounded hash-buckets ($B{=}2^{15}$) prevent vocabulary explosion while preserving hierarchical label order informative for tunneling.
  \item Temporal stability: Exponential-forget gating smooths irregular DNS traffic bursts and improves recall over classic LSTM gates.
  \item Compact compute: About 2.4 M parameters and sub-millisecond latency make the model deployable on commodity hardware.
\end{itemize}

\subsection{System overview}
\begin{figure*}[t]
  \centering
  \includegraphics[width=\textwidth]{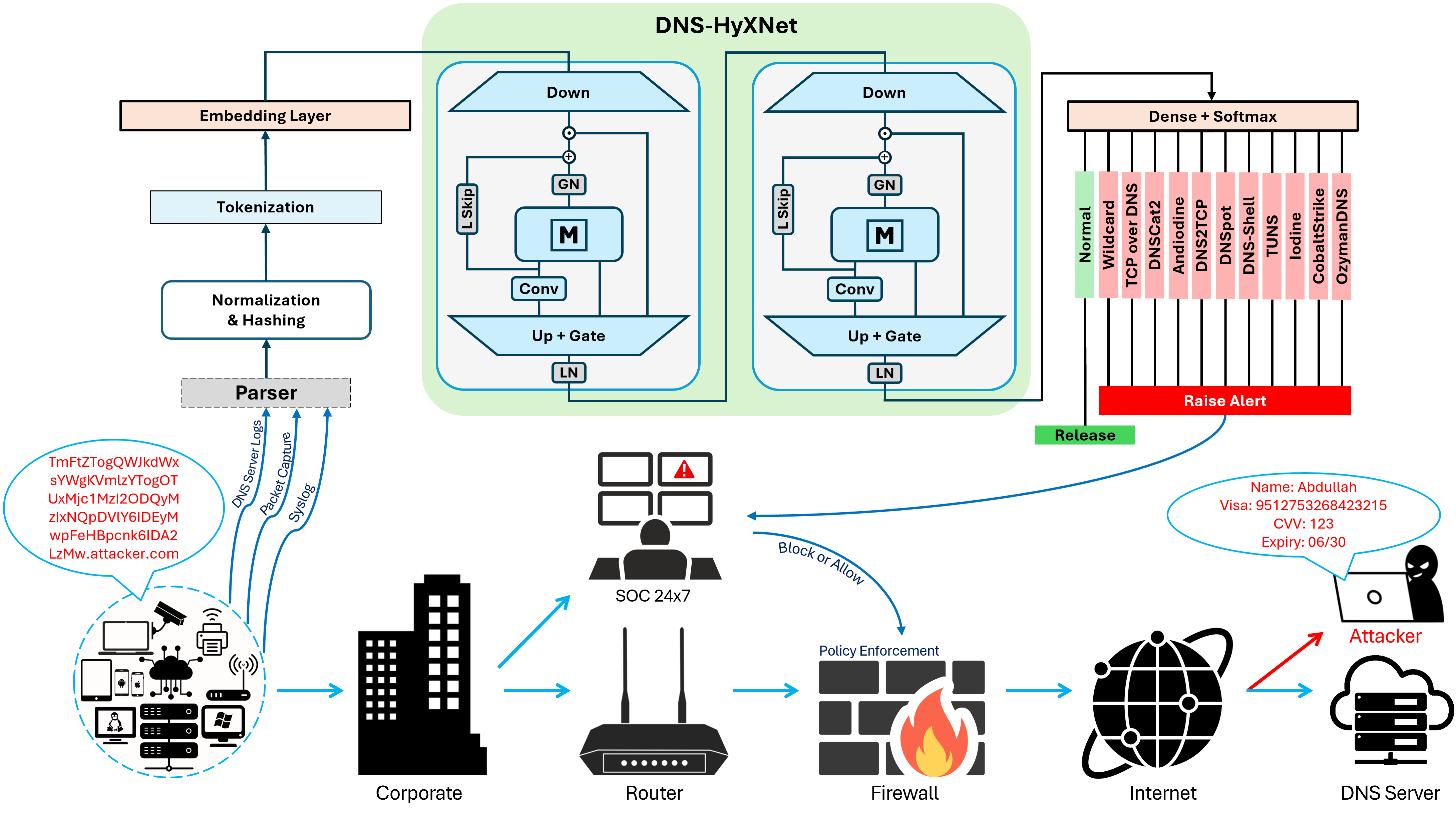}
  \caption{Hybrid xLSTM pipeline and operational integration.
  Raw DNS telemetry from corporate networks (server logs, packet captures, or syslog) is parsed once 
  and converted into compact windowed sequences. 
  The preprocessing layer performs normalization and bounded hash-bucket tokenization, producing 
  numeric and categorical embeddings that feed into two stacked xLSTM blocks. 
  Each block uses exponential forget gating to capture fine-grained temporal dependencies.
  The final dense-softmax head provides a single-stage multi-class decision across benign, wildcard, and tunneling tools. 
  The red arrow at the bottom right indicates the attack flow, showing the adversary's attempt to communicate or exfiltrate data through the DNS channel.
  The model integrates directly with SOC monitoring and policy enforcement systems for real-time blocking or alerting of DNS-based exfiltration attempts.}
  \label{fig:xlstm-architecture}
\end{figure*}

The overall workflow of the proposed framework is illustrated in Figure~\ref{fig:xlstm-architecture}. The system begins by capturing DNS query and response logs through a lightweight parser, which transforms the traffic stream into structured event sequences. These sequences are segmented into fixed-length windows of $T$ events (as described in Section~\ref{sec:windowing}), where each event is represented by two components: (i) a compact vector of numeric metadata (such as packet and record-level statistics), and (ii) a bounded-token encoding of the queried domain name. The token sequences are embedded and processed through two stacked xLSTM layers, whose final hidden representation is fused with the normalized numeric vector. This combined feature vector is then passed to a compact multilayer perceptron (256$\rightarrow$128$\rightarrow K$) with ReLU activations and dropout regularization, yielding a single-stage $K$-way classification output that distinguishes between benign, wildcard, and tunneling traffic families.

\subsection{Problem formulation}
Let $\{p_t\}$ be the DNS packet stream observed at a gateway/resolver. From a sliding window of length $T$, we construct
\[
x_t=\big[\,n_t \,\|\, c_t\,\big],\quad n_t\in\mathbb{R}^{d_n},\; c_t\in\mathbb{N}^{d_c},\quad t=1,\ldots,T,
\]
where $n_t$ are numeric descriptors (e.g., frame.len, dns.resp.ttl, counters), and $c_t$ are categorical tokens derived from the queried name (Section \ref{sec:hashing}). The sequence $\mathcal{S}=\{x_t\}_{t=1}^{T}$ is labeled $y\in\{1,\ldots,K\}$.

\subsection{Streaming-friendly input encoding}\label{sec:hashing}
Efficient real-time detection requires an encoding strategy that is both compact and resilient to the variability of DNS query structures. To achieve this, the proposed framework employs a lightweight combination of numeric normalization and categorical tokenization, beginning with a bounded hashing scheme for domain labels.
\subsubsection{Bounded hash-bucket tokenization}
To prevent vocabulary explosion and keep memory predictable under high-entropy or wildcard domains, we map each domain label (split by dots) into a fixed vocabulary using a 64-bit hash:
\[
\mathrm{bucket}(s)=H(s)\bmod B,\qquad B=2^{15}.
\]
A domain with labels $(t_1,\ldots,t_m)$ becomes $\mathbf{t}(q)=(b_1,\ldots,b_m)$ with $b_j=\mathrm{bucket}(t_j)$. We left-pad to length $T$ with a pad index $0$:
\[
\tilde{\mathbf{t}}(q)=\underbrace{(0,\ldots,0)}_{\max(0,\,T-m)}\circ\mathbf{t}(q),\qquad \tilde{\mathbf{t}}(q)\in\{0,\ldots,B{-}1\}^{T}.
\]
Left-padding preserves right-aligned suffixes (e.g., SLD.TLD) at consistent positions, which empirically stabilizes learning.

\subsubsection{Numeric standardization.}
For numeric features $n_t$, we apply standardization with precomputed $(\mu,\sigma)$ learned on the training split:
\[
\hat{n}_t=(n_t-\mu)\oslash\sigma.
\]
At each step $t$, we build a mixed embedding:
\[
e_t=\big[\,\hat{n}_t \,\|\, \mathrm{Embed}(\tilde{\mathbf{t}}(q_t))\,\big]\in\mathbb{R}^{d_e},
\]
where $\mathrm{Embed}(\cdot)$ is a look-up into an $B\times d_{\text{emb}}$ table with $d_{\text{emb}}=64$.

\subsection{Backbone: xLSTM with exponential-forget gating}
Let $\{e_t\}_{t=1}^T$ be the embedded sequence ($e_t\in\mathbb{R}^{d_{\text{emb}}}$). We use two stacked xLSTM layers with hidden size $h=128$. Each xLSTM cell replaces the classic LSTM forget gate with a continuous exponential decay that improves stability on bursty/irregular traffic.

For input $x_t\in\mathbb{R}^{d}$ and state $(h_{t-1},c_{t-1})\in\mathbb{R}^{h}\times\mathbb{R}^{h}$:
\[
[i_t,f_t,o_t,g_t]=W_xx_t+W_hh_{t-1}+b,
\quad i_t=\sigma(i_t),\; o_t=\sigma(o_t),\; g_t=\tanh(g_t),
\]
\[
\alpha_t=\exp\!\big(-\mathrm{softplus}(f_t)\big)\in(0,1],\quad
c_t=\alpha_t\odot c_{t-1}+i_t\odot g_t,\quad
h_t=o_t\odot\tanh(c_t).
\]
The decay factor $\alpha_t$ yields smooth memory updates (as opposed to hard gating), mitigating gradient pathologies and making the model resilient to long idle gaps followed by sudden domain bursts, common in DNS exfiltration.

\subsection{Feature fusion and single-stage head}
From the top xLSTM layer we take the final hidden state $h_T\in\mathbb{R}^{128}$ and concatenate it with the standardized numeric vector $\hat{x}_{\text{num}}\in\mathbb{R}^{d_n}$:
\[
z=[\,\hat{x}_{\text{num}} \,\|\, h_T\,]\in\mathbb{R}^{d_n+128}.
\]
A compact MLP head performs single-stage multi-class classification:
\begin{align*}
z &\xrightarrow{\;\texttt{Linear}(d_n{+}128,256)\;\to\;\mathrm{ReLU}\;\to\;\mathrm{Dropout}(0.2)\;} u\\
 u &\xrightarrow{\;\texttt{Linear}(256,128)\;\to\;\mathrm{ReLU}\;\to\;\mathrm{Dropout}(0.2)\;} v\\
 v &\xrightarrow{\;\texttt{Linear}(128,K)\;} \ell,
\end{align*}
\[
\hat{y}=\mathrm{softmax}(\ell).
\]
This single head jointly produces maliciousness and tool-family labels, avoiding the two-stage (binary$\rightarrow$tool) pipelines that add latency and maintenance burden.

\subsection{Training objective and optimization}
We train DNS-HyXNet by minimizing the multi-class cross-entropy loss, given by
\begin{equation}
    \mathcal{L} = -\frac{1}{N}\sum_{i=1}^{N} \log \hat{y}_{i,y_i},
\end{equation}
where $N$ is the batch size and $\hat{y}_{i,y_i}$ denotes the predicted probability
assigned to the true class label $y_i$.

We optimize this loss using AdamW (learning rate $2{\times}10^{-3}$, 
weight decay $10^{-4}$). For efficiency, the model employs AMP mixed precision, 
gradient clipping (1.0), a ReduceLROnPlateau scheduler, and early stopping. 
The total parameter count is $\sim$2.4M (64-d embeddings, two 128-hidden xLSTM layers, 
and a compact MLP head), enabling real-time performance on commodity hardware.

\subsection{Windowing and latency budget}\label{sec:windowing}
We fix the token sequence length to $T=15$ steps (left-padded). This choice balances signal capture (enough temporal context to reveal tunneling bursts, wildcard cascades, and C2 cadence) with tight latency. The end-to-end decision latency equals the window fill time plus one forward pass over $T$ steps. Because no recursive graphs or adjacency structures are built, inference cost is dominated by two xLSTM sweeps and a small MLP, typically sub-millisecond per window on a GPU and low-millisecond on a CPU.

\subsection{Online inference logic}
At deployment, incoming packets populate a rolling buffer. Each time $T$ events are available, we (i) tokenize the query labels with the bounded hash, (ii) standardize numeric fields, (iii) perform one xLSTM forward pass, and (iv) apply a threshold to the maximum softmax probability. High-confidence non-benign predictions trigger alerting/blocking; benign traffic is released immediately. Because the encoder is stateless and graph-free, no per-domain or cross-query adjacency must be maintained, simplifying integration with firewalls and SIEM.

\subsection{Complexity discussion}
Let $d_{\text{emb}} = 64$, $h = 128$, and $T = 15$.  
For each window of $T$ events, the recurrent computational cost of a single
xLSTM layer is given by
\begin{equation}
    \mathcal{O}\!\big(T\,(d_{\text{emb}}\cdot 4h + h \cdot 4h)\big)
    = \mathcal{O}\!\big(T\,(256h + 4h^{2})\big).
\end{equation}
In addition, the model includes two small affine layers and a $K$-class
projection. With $h=128$, this results in only a few hundred thousand multiply (accumulate operations per window) two orders of magnitude lower than typical GNN+CNN pipelines that must first construct resolution graphs and then apply graph convolutions.

\subsection{Reproducible configuration}
Unless otherwise stated, we use: two xLSTM layers ($L{=}2$), hidden size $h{=}128$, embedding dimension $d_{\text{emb}}{=}64$, token buckets $B{=}2^{15}$, pad length $T{=}15$, MLP head 256$\rightarrow$128$\rightarrow K$ with ReLU and dropout 0.2, optimizer AdamW (lr $2{\times}10^{-3}$, weight decay $10^{-4}$), AMP mixed precision, gradient clipping 1.0, ReduceLROnPlateau, and early stopping. These match the released code and checkpoint.

\subsection{Contrast to graph-based baselines}
Graph-based detectors reconstruct recursive resolution graphs and apply GNN aggregation (often followed by a second-stage tool classifier). While structurally expressive, they incur (i) non-trivial parsing latency, (ii) sizeable adjacency/state memory, and (iii) multi-stage decision delay. DNS-HyXNet achieves comparable detection capability by learning temporal structure directly from sequences, delivering a single-stage prediction with tight latency and a small, deployment-friendly footprint.

\section{Results and Evaluation}\label{sec4}
This section presents the empirical assessment of DNS-HyXNet, including the experimental setup, dataset characteristics, comparative baselines, and quantitative performance across multiple evaluation scenarios. We begin by detailing the experimental configuration used throughout the study.
\subsection{Experimental Setup}
All experiments were performed on two publicly available benchmarks that are widely adopted in DNS tunnel research.  
The first, DNS-Tunnel-Datasets, was introduced by Gao et~al. in the GraphTunnel study~\cite{gao2024graphtunnel}.  
It contains benign, wildcard, and eleven tunneling families generated from popular DNS tunneling tools, including dnscat2, iodine, dns2tcp, tuns, and CobaltStrike.  
The second dataset, CIC-Bell-DNS-EXF-2021~\cite{cicdns2021}, released by the Canadian Institute for Cybersecurity, provides binary labels distinguishing exfiltration from legitimate DNS traffic.  
Both datasets were processed using the unified preprocessing and feature pipeline described in Section~\ref{sec3}.  
Experiments were conducted separately for each dataset to preserve comparability.

Training, validation, and testing partitions followed variable splits to assess robustness under differing data availability.  
Training ratios were varied from 60\% to 10\%, with 20\% fixed for validation and the remainder for testing.  
All models were optimized using the AdamW optimizer with mixed-precision (AMP) training for computational efficiency.  
Early stopping was employed when validation loss converged.  
The implementation was executed on a Lenovo Legion 5 computing system featuring an NVIDIA RTX 4050 GPU with CUDA support.
Inference ran without runtime bottlenecks, confirming the efficiency and streaming capability of the model.

\subsection{Datasets Overview}
\begin{figure}[!t]
  \centering
  \begin{subfigure}[t]{0.51\columnwidth}
    \centering
    \includegraphics[width=\linewidth]{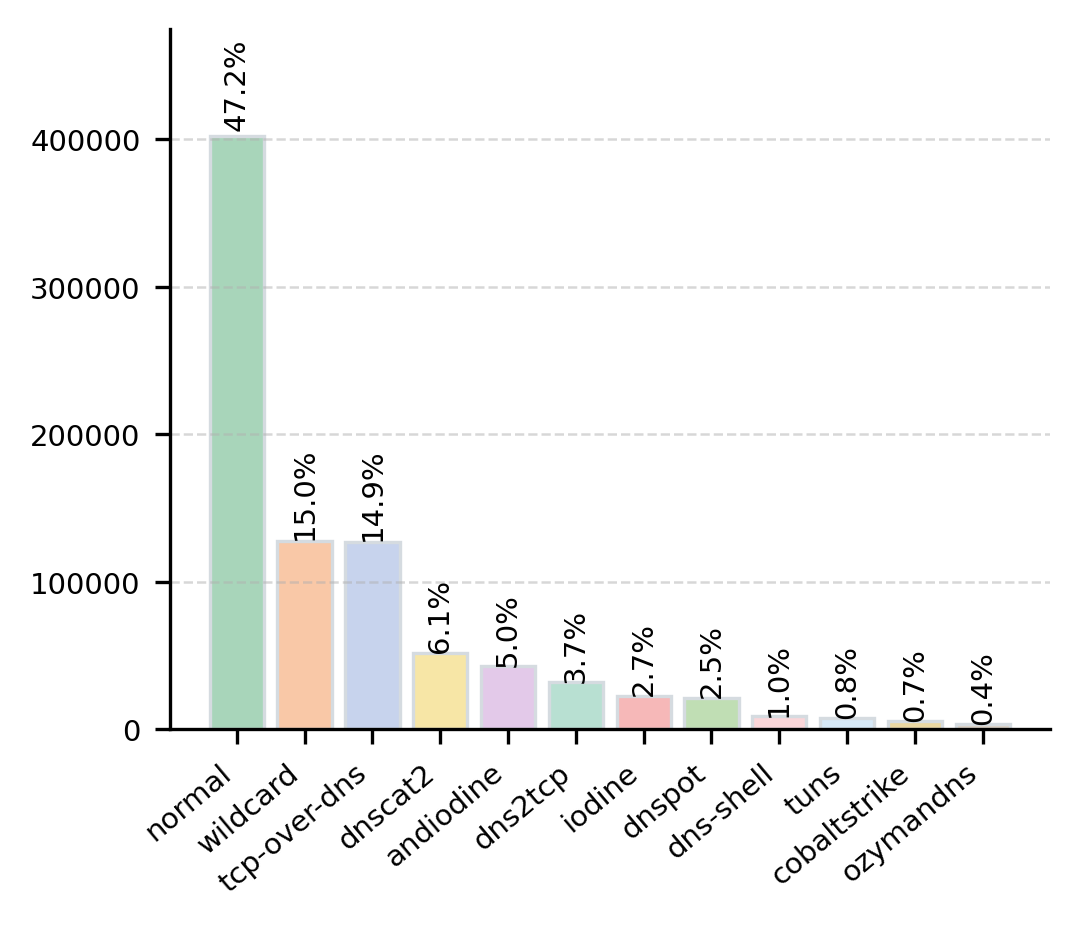}
    \caption{Class-wise distribution in the DNS-Tunnel-Datasets.}
    \label{fig:dsclass}
  \end{subfigure}\hfill
  \begin{subfigure}[t]{0.45\columnwidth}
    \centering
    \includegraphics[width=\linewidth]{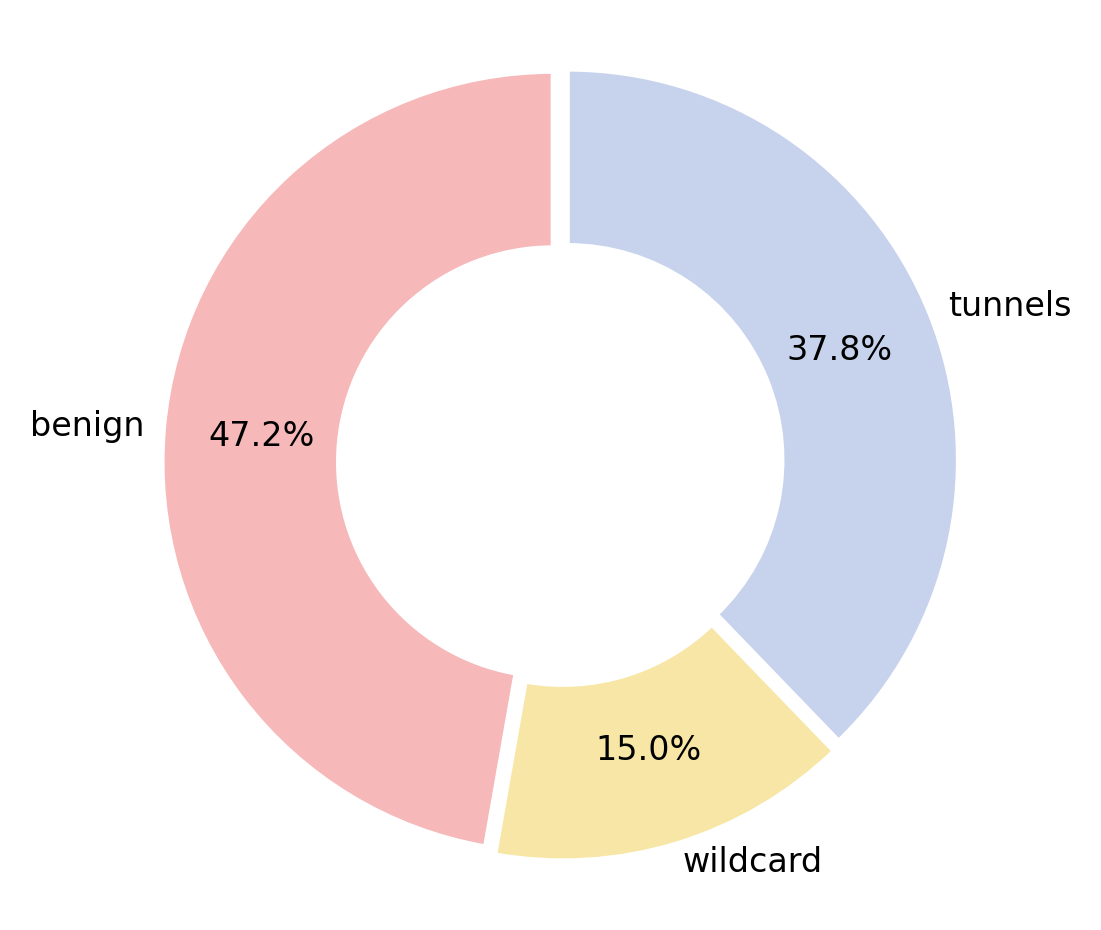}
    \caption{Traffic composition of benign, wildcard, and tunneling DNS in the dataset.}
    \label{fig:pie}
  \end{subfigure}
  \caption{Overall visualization of dataset characteristics in the DNS-Tunnel-Datasets.}
  \label{fig:ds_distributions}
\end{figure}

The per-class distribution of the DNS-Tunnel datasets is shown in Figure \ref{fig:dsclass}.  
As expected, benign and wildcard queries dominate the data, followed by tunneling families with smaller sample counts such as dns-shell, tuns, and ozymandns.
Despite this imbalance, DNS-HyXNet maintained consistent detection accuracy across all classes.  
The overall composition of benign, wildcard, and tunneling traffic (where tunneling represents approximately 38\% of total samples) is summarized in Figure \ref{fig:pie}.

\begin{figure}[!b]
\centering
\includegraphics[width=\columnwidth]{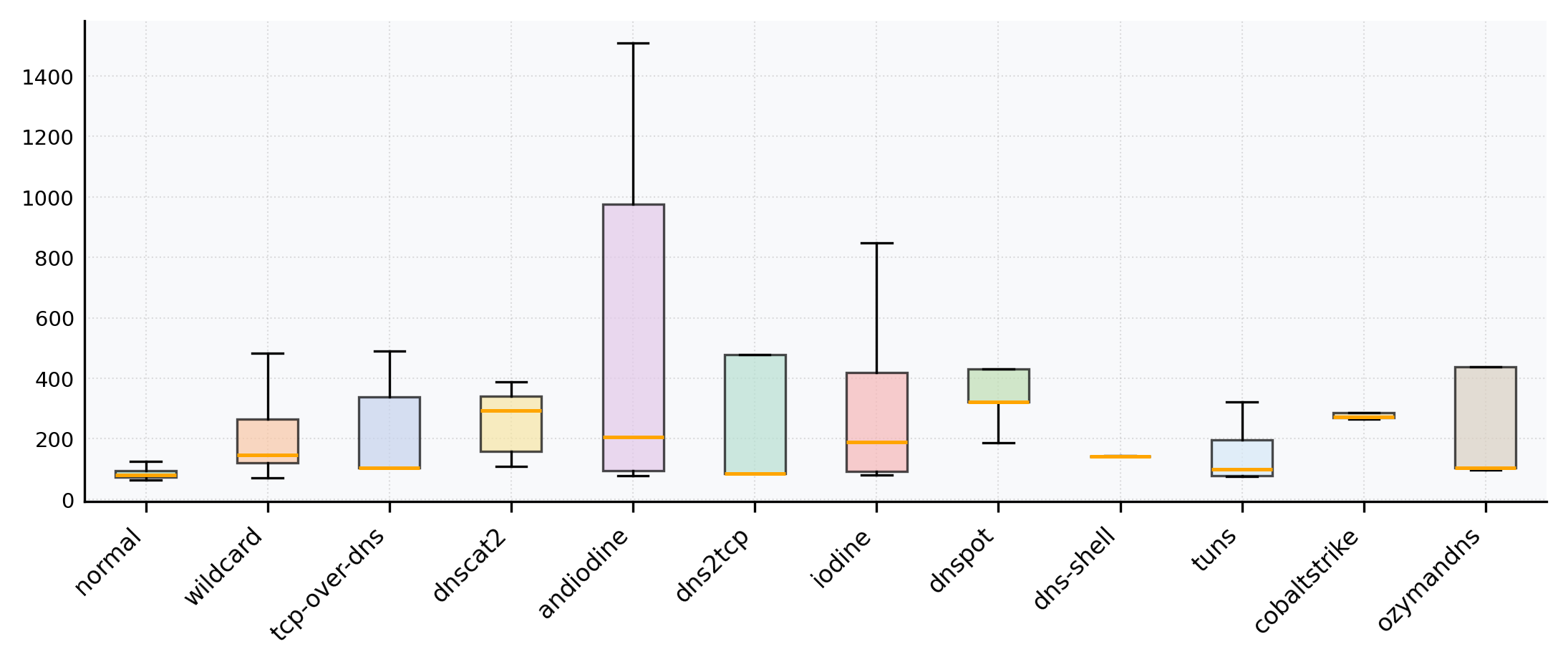}
\caption{Distribution of frame.len values across DNS traffic classes. Tunnel traffic exhibits greater length variability compared to normal DNS queries.}
\label{fig:boxlen}
\end{figure}

In addition to categorical imbalance, notable numerical disparities were observed across low-level packet features.  
For instance, frame.len (packet length) varies significantly between tunneling and benign traffic, while TTL-related parameters show characteristic distributions unique to each tunneling tool.  
Figure~\ref{fig:boxlen} visualizes the spread of frame.len values per class, indicating longer and more variable frames for tunneling tools such as andiodine and ozymandns.  

\subsection{Performance on the DNS-Tunnel-Datasets}

DNS-HyXNet demonstrates consistently strong performance across all evaluation settings of the DNS-Tunnel-Datasets. Even under reduced training data availability, the model maintained near-perfect predictive accuracy, reflecting its ability to generalize from limited supervision. This stability is further reflected in the per-class evaluation results, which show uniformly high precision and recall across all twelve traffic categories.

\begin{figure}[!t]
\centering
\includegraphics[width=0.9\textwidth]{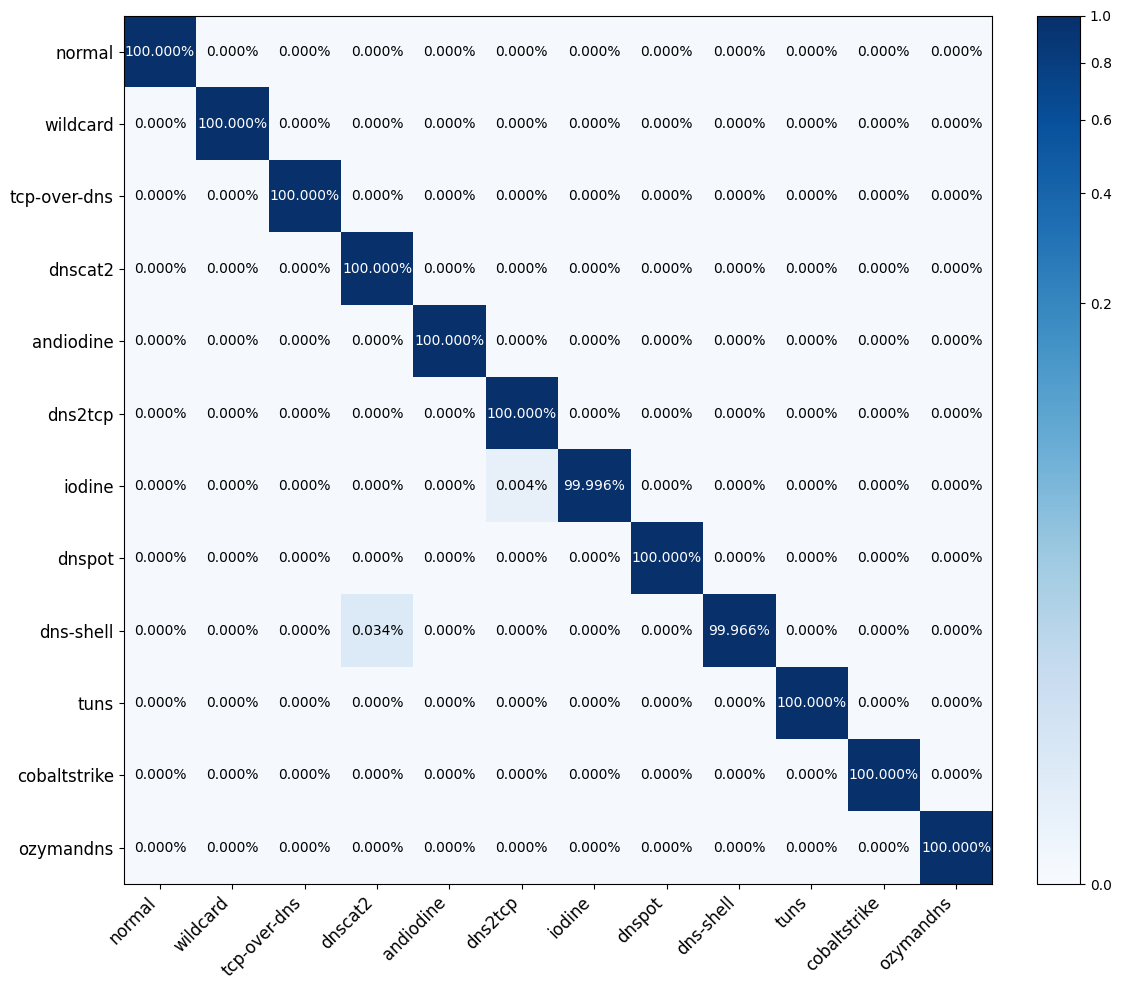}
\caption{Row-normalized confusion matrix for the DNS-Tunnel-Datasets (60 - 20 - 20 split). Off-diagonal elements are negligible.}
\label{fig:cm_dns}
\end{figure}

Per-class performance remained uniformly excellent.  
All traffic categories (normal, wildcard, tcp-over-dns, dnscat2, andiodine, dns2tcp, iodine, dnspot, dns-shell, tuns, CobaltStrike, and OzymanDNS) achieved precision and recall values exceeding 0.999.  
Figure~\ref{fig:cm_dns} shows the row-normalized confusion matrix for the 60–20–20 split, where the off-diagonal mass is essentially zero, confirming that misclassifications were negligible.

Compared to GraphTunnel, which reports a detection F1-score of 0.9978 in its most challenging wildcard-DNS experiments on the same DNS-Tunnel-Datasets
\cite{gao2024graphtunnel}, DNS-HyXNet attains an F1-score of approximately 0.9999 while eliminating the need for recursive graph reconstruction.
This demonstrates that temporal sequence modeling via xLSTM captures the salient behavioral structure of DNS tunneling activity without relying on explicit graph topology, yielding comparable or better accuracy with significantly reduced computational overhead.

\subsection{Performance on CIC-Bell-DNS-EXF-2021 dataset}

\begin{table}[!b]
\centering
\caption{Comparison of DNS-HyXNet with published methods on the CIC-Bell-DNS-EXF-2021 dataset.
Red arrows ($\uparrow$) indicate a higher value than DNS-HyXNet; blue arrows ($\downarrow$) indicate a lower value.}
\label{tab:cic_compare}
\footnotesize
\renewcommand{\arraystretch}{1.15}
\setlength{\tabcolsep}{4pt}

\begin{tabularx}{\columnwidth}{
    L{3.2cm}
    C{2.2cm}
    C{2.2cm}
    C{2.2cm}
    C{2.2cm}}
\toprule
\textbf{Method} & \textbf{Accuracy} & \textbf{Precision} &
\textbf{Recall} & \textbf{F1-score} \\
\midrule

Samaneh et al.~\cite{samaneh2020dns} 
& 0.9997 {\color{red}($\uparrow$0.0026)} 
& 0.9997 {\color{red}($\uparrow$0.0022)}
& 0.9997 {\color{red}($\uparrow$0.0034)}
& 0.9997 {\color{red}($\uparrow$0.0028)} \\

Suman et al.~\cite{suman2021dns}
& 0.989 {\color{blue}($\downarrow$0.0081)}
& 0.992 {\color{blue}($\downarrow$0.0055)}
& 0.989 {\color{blue}($\downarrow$0.0073)}
& 0.989 {\color{blue}($\downarrow$0.0079)} \\

Filippo et al.~\cite{filippo2019dns}
& 0.971 {\color{blue}($\downarrow$0.0261)}
& 0.945 {\color{blue}($\downarrow$0.0525)}
& 0.956 {\color{blue}($\downarrow$0.0403)}
& 0.956 {\color{blue}($\downarrow$0.0409)} \\

D’Angelo et al.~\cite{dangelo2020dns}
& 0.9977 {\color{red}($\uparrow$0.0006)}
& 0.9995 {\color{red}($\uparrow$0.0020)}
& 0.9971 {\color{red}($\uparrow$0.0008)}
& 0.9971 {\color{red}($\uparrow$0.0002)} \\

GraphTunnel~\cite{gao2024graphtunnel}
& 1.0000 {\color{red}($\uparrow$0.0029)} 
& 1.0000 {\color{red}($\uparrow$0.0025)} 
& 1.0000 {\color{red}($\uparrow$0.0037)} 
& 1.0000 {\color{red}($\uparrow$0.0031)} \\

\midrule
\textbf{DNS-HyXNet}
& \textbf{0.9971} 
& \textbf{0.9975}
& \textbf{0.9963}
& \textbf{0.9969} \\
\bottomrule
\end{tabularx}
\end{table}



\begin{figure}[!t]
  \centering
  \begin{subfigure}[t]{0.40\columnwidth}
    \centering
    \includegraphics[width=\linewidth]{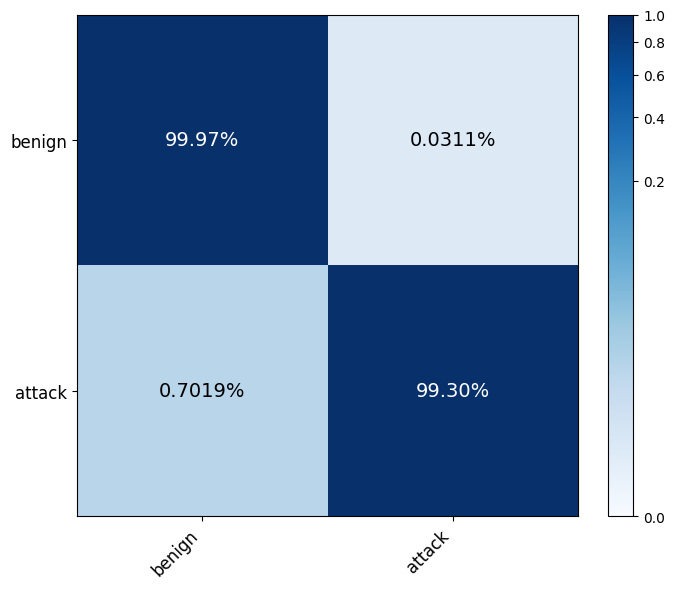}
    \caption{Row-normalized confusion matrix for CIC-Bell-DNS-EXF-2021.}
    \label{fig:cm_cic}
  \end{subfigure}\hfill
  \begin{subfigure}[t]{0.56\columnwidth}
    \centering
    \includegraphics[width=\linewidth]{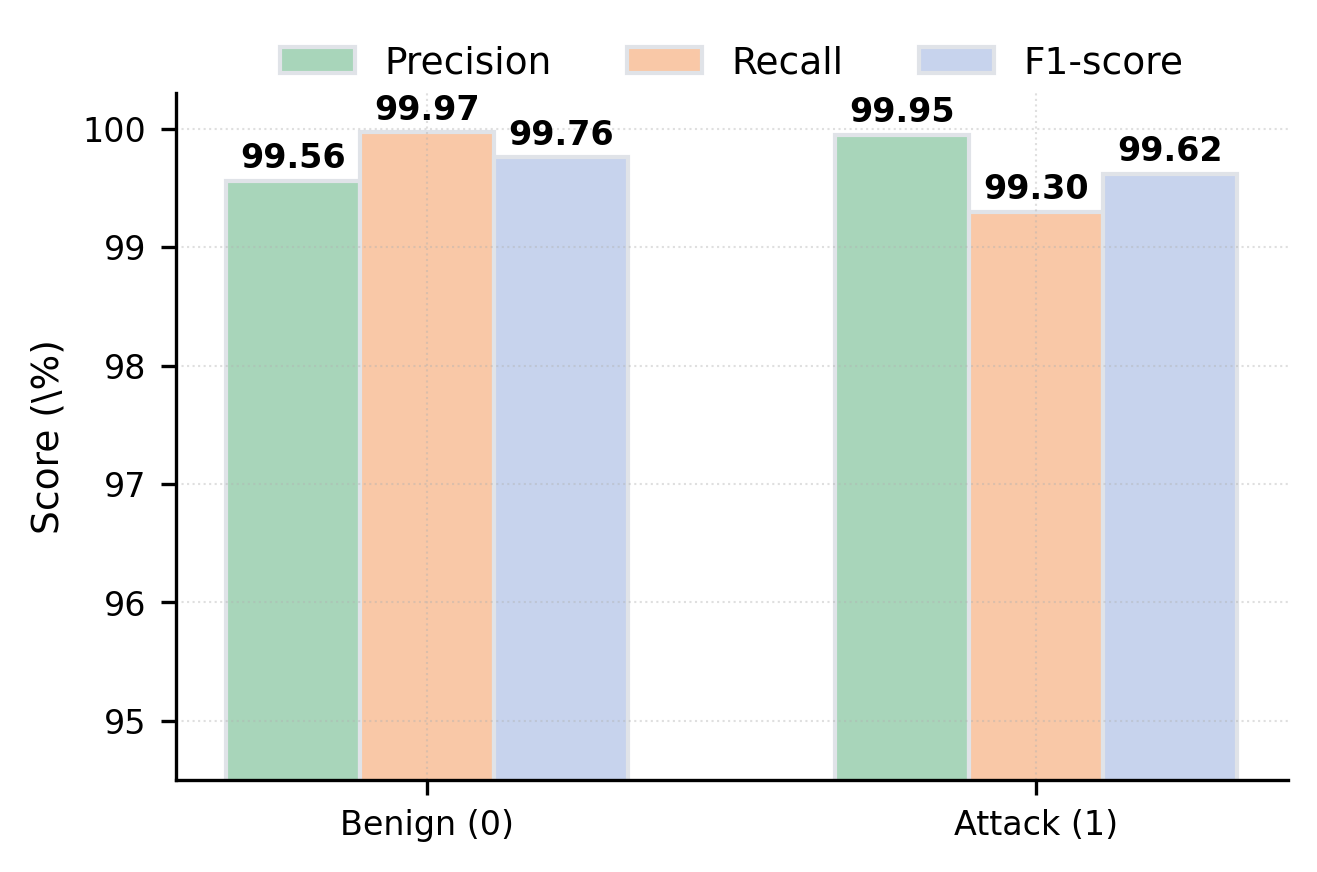}
    \caption{Per-class precision, recall, and F1-score.}
    \label{fig:cicbars}
  \end{subfigure}
  \caption{Overall performance of DNS-HyXNet on the CIC-Bell-DNS-EXF-2021 dataset.  
  Subfigure~\subref{fig:cm_cic} shows the row-normalized confusion matrix, 
  while Subfigure~\subref{fig:cicbars} highlights that both classes achieve precision, recall,
  and F1-scores above 99.3\%.}
  \label{fig:cic_overall}
\end{figure}

To further validate generalization on independent data, DNS-HyXNet was tested on the CIC-Bell-DNS-EXF-2021 dataset. The model achieved an overall accuracy of 99.71\%, corresponding to macro-averaged precision, recall, and F1-scores of 99.75\%, 99.63\%, and 99.69\%, respectively. Only 61 misclassifications occurred across 20{,}977 samples.
Table~\ref{tab:cic_compare} compares these aggregates with previously published methods: DNS-HyXNet is competitive with the strongest deep-learning baselines and clearly surpasses earlier machine-learning approaches, while GraphTunnel attains slightly higher scores (1.000 across all four metrics) at the cost of graph construction and a more complex two-stage inference pipeline. Figure~\ref{fig:cic_overall} summarizes our per-class behavior: the confusion matrix in Subfigure~\subref{fig:cm_cic} shows only 61 misclassifications, and the bar plot in Subfigure~\subref{fig:cicbars} confirms that both benign and attack classes achieve precision, recall, and F1-scores above 99.3\%.


\begin{table}[!t]
\centering
\caption{Comparison of DNS-HyXNet with representative DNS tunneling detection baselines.\newline
Only DNS-HyXNet and GraphTunnel report detection performance on the DNS-Tunnel-Datasets;
for the remaining methods, Accuracy and F1 are marked as N/A because they are evaluated on different corpora.\newline
\footnotesize\textit{Columns:} Graph req. = Graph construction required; Real-time = real-time capability. \newline
\textit{Symbols:} \cmark~supported, \xmark~not supported, \pmark~partially or unclearly reported.}
\label{tab:compare_baseline}
\footnotesize
\renewcommand{\arraystretch}{1.18}
\setlength{\tabcolsep}{3pt}

\begin{tabularx}{\columnwidth}{
  L{2.2cm}
  X
  C{1.0cm}
  C{2.0cm}
  C{1.9cm}
  C{1.1cm}
  C{1.1cm}}
\toprule
\textbf{Method} & \textbf{Architecture} & \textbf{Stages} &
\textbf{Accuracy ({\color{blue}↓}{\color{red}↑})} &
\textbf{F1 ({\color{blue}↓}{\color{red}↑})} &
\textbf{Graph req.} &
\textbf{Real-time} \\
\midrule

GraphTunnel~\cite{gao2024graphtunnel}
& GNN + CNN 
& Two
& 0.9978 {\color{blue}(--0.0021$\downarrow$)}
& 0.9978 {\color{blue}(--0.0021$\downarrow$)}
& \cmark & \xmark \\

FECC~\cite{liang2023fecc}
& CNN + Clustering
& One
& N/A
& N/A
& \xmark & \pmark \\

DNS-Images~\cite{d2022dns}
& CNN (2D image)
& One
& N/A
& N/A
& \xmark & \pmark \\

Rule-based~\cite{adiwal2023dns}
& Heuristic
& One
& N/A
& N/A
& \xmark & \cmark \\

\midrule
\textbf{DNS-HyXNet}
& \textbf{Sequential}
& \textbf{One}
& \textbf{0.9999}
& \textbf{0.9999}
& \textbf{\xmark}
& \textbf{\cmark} \\
\bottomrule
\end{tabularx}
\end{table}







%


\subsection{Comparative Analysis}
A comparative summary of DNS-HyXNet and related baseline models is presented in Table~\ref{tab:compare_baseline}.
GraphTunnel’s graph neural network achieved strong detection accuracy but incurred significant latency and preprocessing overhead due to recursive graph construction.
CNN-based systems such as FECC~\cite{liang2023fecc} and DNS-Images~\cite{d2022dns} depend on 2D feature transformations, which further increase computational cost.

In contrast, DNS-HyXNet attains competitive detection performance while using a single-stage recurrent architecture that eliminates both recursive graph generation and image-based preprocessing.

\begin{figure}[!b]
\centering

\begin{subfigure}[t]{0.48\columnwidth}
  \centering
  \includegraphics[width=\linewidth]{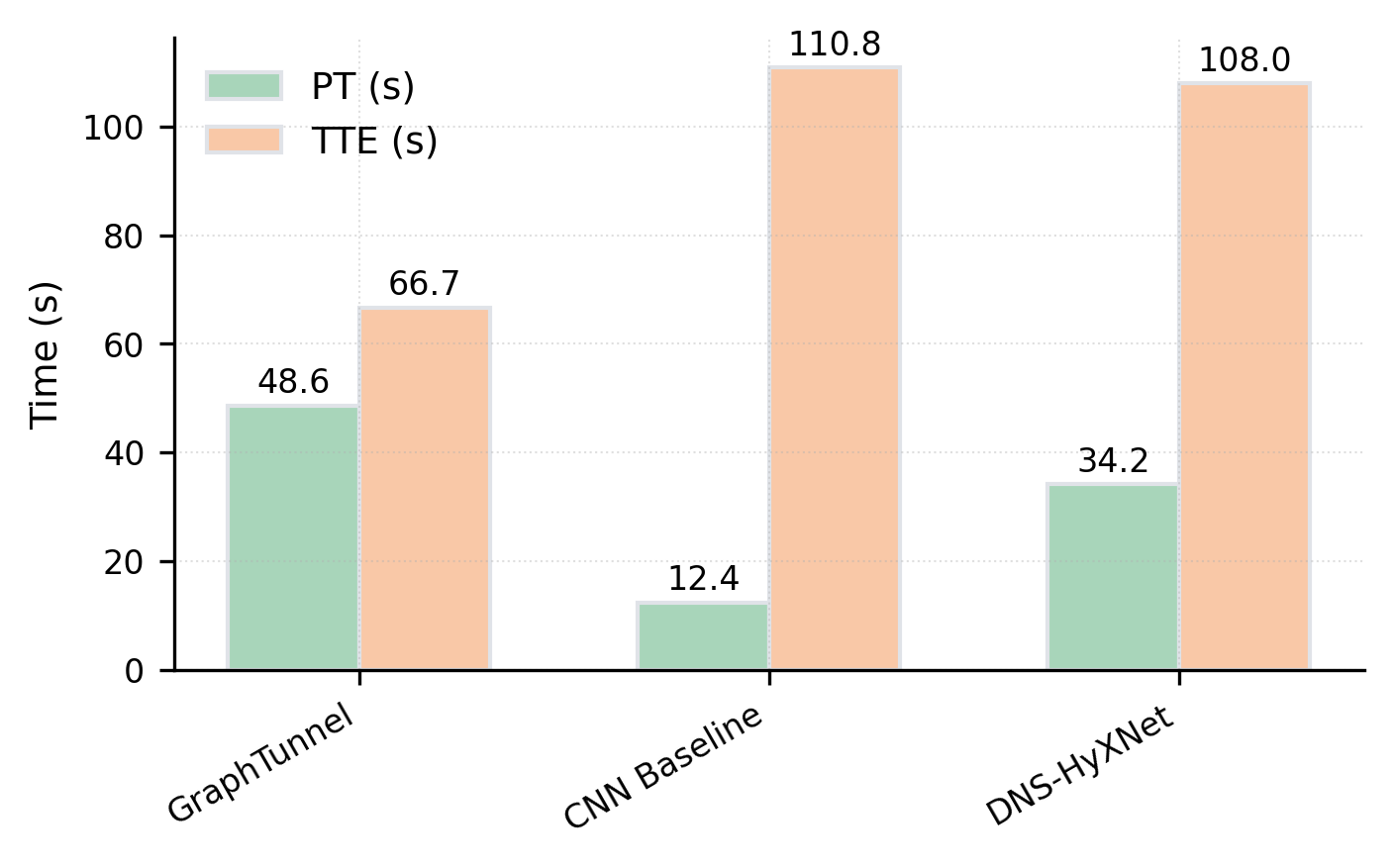}
  \caption{Preprocessing (PT) and total training time (TTE).}
  \label{fig:runtime_pttte}
\end{subfigure}
\hfill
\begin{subfigure}[t]{0.48\columnwidth}
  \centering
  \includegraphics[width=\linewidth]{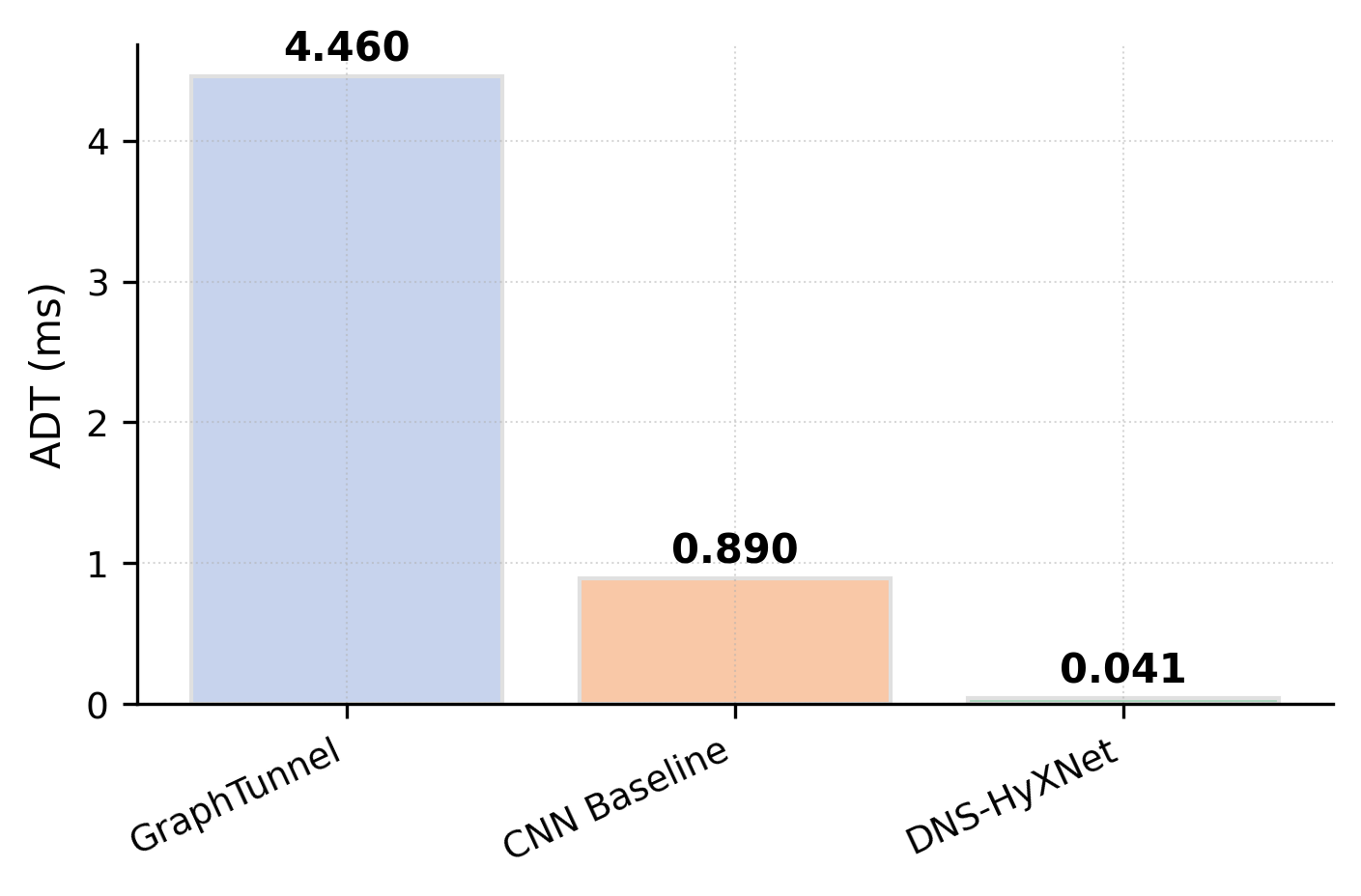}
  \caption{Average Detection Time (ADT) per sample.}
  \label{fig:runtime_adt}
\end{subfigure}

\caption{Runtime profile comparison. DNS-HyXNet maintains competitive preprocessing and training times while achieving the lowest detection latency.}
\label{fig:runtime_profile}
\end{figure}

\begin{table}[!t]
\centering
\caption{Runtime and resource comparison of DNS-HyXNet (measured programmatically).\newline
\footnotesize\textit{Columns:} PT = Preprocessing time; TTE = Total training time; 
ADT = Average detection time per sample; Throughput = Samples per second; 
GPU peak = Peak GPU memory usage.}
\label{tab:runtime}
\footnotesize
\renewcommand{\arraystretch}{1.08}
\setlength{\tabcolsep}{3pt}
\begin{tabularx}{\columnwidth}{
    >{\raggedright\arraybackslash}p{2.3cm}
    >{\centering\arraybackslash}p{2.0cm}
    >{\centering\arraybackslash}p{2.0cm}
    >{\centering\arraybackslash}p{1.8cm}
    >{\centering\arraybackslash}p{1.9cm}
    >{\centering\arraybackslash}p{1.9cm}
}
\toprule
Model & PT (s)({\color{blue}↓}{\color{red}↑}) & TTE (s)({\color{blue}↓}{\color{red}↑}) & ADT (ms)({\color{blue}↓}{\color{red}↑}) & 
Throughput (samples/s) & GPU peak (MB)({\color{blue}↓}{\color{red}↑})\\
\midrule
GraphTunnel~\cite{gao2024graphtunnel} 
  & 48.65 {\color{red}(\,14.44\,$\uparrow$\,)}     
  & 66.69 {\color{blue}(\,$-41.3$\,$\downarrow$\,)} 
  & 4.46  {\color{red}(\,4.42\,$\uparrow$\,)}       
  & -- 
  & 126.4 {\color{blue}(\,$-815$\,$\downarrow$\,)}  
\\

CNN Baseline~\cite{d2022dns} 
  & 12.39 {\color{blue}(\,$-21.8$\,$\downarrow$\,)} 
  & 110.85 {\color{red}(\,2.85\,$\uparrow$\,)}      
  & 0.89  {\color{red}(\,0.85\,$\uparrow$\,)}       
  & -- 
  & 1810.9 {\color{red}(\,869\,$\uparrow$\,)}       
\\
\specialrule{0.11em}{0.2em}{0.2em}

\textbf{DNS-HyXNet} 
  & \textbf{34.21} 
  & \textbf{108.0}$^\dagger$ 
  & \textbf{0.041} 
  & \textbf{24,215} 
  & \textbf{941.7} \\
\bottomrule

\multicolumn{6}{l}{\footnotesize{$^\dagger$Per-epoch equivalent time; full training $\approx$ 2154 s.}}
\end{tabularx}
\end{table}

\subsection{Runtime and Efficiency}
To evaluate operational scalability, runtime and memory consumption were profiled programmatically.
Table~\ref{tab:runtime} reports preprocessing time (PT), total training time (TTE), average detection time (ADT), and throughput for DNS-HyXNet in comparison with GraphTunnel and a CNN baseline.
Although the total training time of DNS-HyXNet reflects full multi-epoch execution, its per-epoch duration remains comparable to both CNN and GNN baselines.
Most importantly, DNS-HyXNet achieves an average detection time of only 0.041~ms per sample (equivalent to more than 24k samples processed per second) demonstrating clear real-time readiness.
Note that Table~\ref{tab:runtime} reports only GPU peak memory; GraphTunnel also maintains additional graph adjacency state on the CPU side. In contrast, DNS-HyXNet stores only a fixed-length sequence buffer, yielding a predictable overall memory footprint even as the number of active domains grows.
Figure~\ref{fig:runtime_profile} further visualizes these runtime characteristics.
GraphTunnel exhibits substantial preprocessing overhead and higher detection latency,
while the CNN baseline is constrained by heavy computational and memory demands.
In contrast, DNS-HyXNet provides a favorable balance: competitive training efficiency
combined with the lowest detection latency of all evaluated models, making it the most
practical solution for real-time DNS tunnel detection.

\section{Discussions}\label{sec5}
The experimental results confirm that DNS-HyXNet effectively bridges the performance gap between computationally intensive graph-based models and lightweight, deployable solutions.  
Unlike GraphTunnel~\cite{gao2024graphtunnel}, which relies on recursive resolution graphs and multi-stage inference, DNS-HyXNet attains competitive accuracy using a purely sequential representation of DNS flows.
By leveraging extended LSTM cells with exponential-forget gating, the model captures long-term temporal dependencies within DNS query-response streams, thereby encoding contextual behavior implicitly rather than reconstructing it explicitly through graph topology.

The findings underscore that covert DNS tunnels are better characterized by their temporal dynamics (patterns of query timing, size, and entropy) than by static recursive relationships.  
This temporal abstraction simplifies feature extraction and dramatically lowers runtime complexity: DNS-HyXNet performs single-pass inference with an average detection latency of only 0.041~ms per sample, achieving throughput above 24k samples/s.  
The system therefore satisfies operational requirements for inline traffic monitoring at enterprise gateways or network edges.

The xLSTM-based framework can extend naturally to other encrypted or covert channels, such as DNS over HTTPS (DoH) and DNS over TLS (DoT), where sequential packet behavior remains an observable indicator even under payload encryption.  
Its compact design and low memory footprint (under 1~GB GPU peak) make it suitable for deployment on SOC appliances, IoT gateways, and federated monitoring nodes where scalability and interpretability are crucial.

While current experiments demonstrate outstanding generalization on publicly available datasets, the evaluation is limited to controlled tunneling scenarios and predefined traffic distributions.  
Real-world deployment may involve unseen resolvers, NAT aggregation, or adversarial manipulation of timing and length features.  
Future research will investigate adaptive thresholding, domain generalization, and interpretability techniques (such as temporal attention or saliency mapping) to elucidate decision pathways and sustain accuracy in dynamic network environments.

\section{Conclusions}\label{sec6}

This research presented DNS-HyXNet, an extended LSTM based framework for real-time detection and attribution of DNS tunneling attacks.  
By reformulating tunnel detection as a temporal sequence modeling problem rather than a graph-reconstruction task, the proposed architecture achieves near state-of-the-art detection accuracy while maintaining extremely low computational cost and latency.

Across multiple train–validation–test splits on the DNS-Tunnel-Datasets, DNS-HyXNet consistently maintained accuracy near 99.99\% and macro-averaged precision, recall, and F1-scores above 99.96\%. On the independent CIC-Bell-DNS-EXF-2021 dataset, it maintained 99.71\% accuracy with only 61 misclassifications, confirming its robustness across heterogeneous traffic.  
Runtime profiling revealed per-sample inference latency of 0.041~ms and GPU memory utilization under 1~GB, validating the model’s suitability for deployment in real-time environments.

In contrast to multi-stage or graph-based systems such as GraphTunnel, DNS-HyXNet performs end-to-end multi-class classification in a single pass, eliminating the need for recursive graph parsing or image-based feature transformations.  
The results substantiate that temporal sequence learning alone can effectively capture behavioral signatures of DNS tunneling and exfiltration, marking a paradigm shift toward graph-free, streaming-first intrusion detection.

Looking ahead, the framework will be extended to encrypted DNS and other protocol-level exfiltration vectors.  
Ongoing work will explore online learning, federated adaptation, and interpretable attention mechanisms to ensure that DNS-HyXNet remains transparent, adaptable, and resilient in evolving network ecosystems.

\section*{Data availability}
All datasets analyzed in this study are publicly available.  
The DNS-Tunnel-Datasets introduced by Gao et al.~\cite{gao2024dataset} were used for multi-class DNS tunnel detection experiments,  
while the CIC-Bell-DNS-EXF-2021 dataset~\cite{cicdns2021} was employed for binary exfiltration detection.  
Detailed preprocessing steps, feature construction, and split configurations are described in Section~\ref{sec3} and the Supplementary Information.

\section*{Code availability}
The complete implementation of DNS-HyXNet will be released under an open-source license upon publication to facilitate transparency, reproducibility, and future research.

\section*{Competing interests}
The authors declare no competing interests.

\backmatter
\phantomsection

\bmhead{Acknowledgements}
The authors gratefully acknowledge the support of King Fahd University of Petroleum and Minerals (KFUPM), Dhahran, Saudi Arabia,  
for providing research facilities and computational resources that enabled this study.

\bibliography{sn-bibliography}
\end{document}